\documentclass[a4paper]{article}
\usepackage{onecolpceurws}
\usepackage{graphicx}
\usepackage{hyperref}
\usepackage{tikz}
\usepackage{pgfmath}
\usepackage{booktabs}
\usepackage{caption}

\title{Mind the Gap: a Well Log Data Analysis}

 \author{
 Rui L. Lopes$^{1,2}$ \\ rmlc@inesctec.pt
 \and
 Al\'ipio M. Jorge$^{1,3}$ \\ amjorge@fc.up.pt
 }
   
 \institution{
 	$^1$ LIAAD, INESC TEC, Porto, Portugal\\
 	$^2$ HASLab, INESC TEC, Porto, Portugal\\
    $^3$ DCC - FCUP, Universidade do Porto, Portugal
 }

\begin{document}
\maketitle

\begin{abstract}
The main task in oil and gas exploration is to gain an understanding of the distribution and nature of rocks and fluids in the subsurface. 
Well logs are records of petro-physical data acquired along a borehole, providing direct information about what is in the subsurface. The data collected by logging wells can have significant economic consequences, 
due to the costs inherent to drilling wells, and the potential return of oil deposits.

In this paper, we describe preliminary work aimed at building a general framework for well log prediction.
First, we perform a descriptive and exploratory analysis of the gaps in the neutron porosity logs of more than a thousand wells in the North Sea. Then, we generate artificial gaps in the neutron logs that reflect the statistics collected before. Finally, we compare Artificial Neural Networks, Random Forests, and three algorithms of Linear Regression in the prediction of missing gaps on a well-by-well basis.

\end{abstract}
\vskip 32pt

\section{Introduction}\label{sec:intro}

Understanding the geometry and composition of rocks and fluids in the subsurface is the primary task in the search for oil and gas. The acquired information has a crucial importance for the success and the efficiency of the oil and gas exploration and production activities.
Well logs provide direct information about what is in the subsurface. The data collected by logging wells can have significant economic consequences, first because it has a direct impact on the following drilling decisions, and second due to the costs of drilling, and the potential return of the oil deposits.

Several logs are typically collected from each well. The data readings are usually acquired through sensing tools that are lowered into the hole by cable. The data acquired by each sensor is “logged” (recorded) at the surface as the tool is pulled up the hole. During drilling, data may also be acquired by instrumentation at the bottom of the hole. The term "well log data" is then used to refer to data collected in or descriptive of the rock and fluid surrounding the hole.

These detailed and direct measurements of rock and fluid properties in the subsurface typically include i) gamma ray intensity (related to the types of minerals present); ii) electrical resistance, (related to the quantity and types of fluids); iii) density and porosity (related to the pore-volume fraction); and iv) sonic velocity 
(related to both rock and fluid properties). These are common examples,  illustrative of the hundreds of well logs that may be collected.

However, gaps are sometimes present in well logs. Well log gaps result in less information on which to base a model and consequently more uncertainty regarding what will be encountered when the next well is drilled. Gaps happen for different reasons, including tools that fail or malfunction, or operators that may turn off recording equipment at the wrong time. Moreover, it may be discovered later that the wrong interval was logged.

In this work we first perform a descriptive and exploratory analysis of the gaps in the logs of 1026 wells in the North Sea. The goal of this analysis is to understand the frequency and size of the gaps accross data from hundreds of wells. Having this knowledge it is possible to generate artificial gaps in the neutron logs that reflect the statistics collected before and thus constitute a relevant synthetic problem. 
The usual approach to the prediction of log values uses only data from the same well (or a few wells from the same block), typically surrounding a particular gap \cite{aifa_neuro-fuzzy_2014,zerrouki_prediction_2014,mohseni_application_2015}. Different techniques can be used to predict the missing data. In this work, we compare Artificial Neural Networks, Random Forests, and Generalised Linear Models, Bayesian Regression, and RANdom SAmple Consensus (RANSAC) in the prediction of missing gaps on a well-by-well basis. In particular, we generate artificial gaps on the neutron porosity readings a of a complete well, and train models to predict those values from the remaining sensor data.

The remaining of the article is organized as follows. 
Section \ref{sec:dataset} provides details about the data-set that was used. 
Then, in Section \ref{sec:methodology} the methodology is described, the used algorithms are briefly presented, and the results of the analysis and different models are discussed.
Finally, Section \ref{sec:future} concludes with a discussion on the next steps and ideas for future work.

\section{Data-set and Gaps}\label{sec:dataset}

The characterization of the subsurface relies on detailed and direct information about rock and fluid properties, which is collected by different sensors in wells. 
Typically several logs are collected from each well, acquired by lowering sensing tools into the hole by cable. The data acquired by each sensor is recorded either at the surface as the tool is pulled up the hole, or by instrumentation at the bottom of the hole while drilling is in process. 

The data collected from the various sensors at some particular well is usually merged into a composite log. This log includes the meta-data for the well, and the readings for each sensor aligned by depth (using hard-coded out-of-range values for the missing values). The data is typically reported in a graphical format with the sensor readings side by side for easy visual inspection.
We collected the composite logs from Dutch wells in the North Sea, summing up to a total of around six hundred thousand samples from 1026 wells\footnote{The data and meta-data for each well can be consulted at \url{http://www.nlog.nl/en/listing-boreholes}.}. The logs are furnished by the Dutch government and are freely available at the NLOG site, which is managed by TNO, Geological Survey of the Netherlands.  

The most common logs found in these records were: i) \emph{bulk density}, which relates to the seismic velocity of waves travelling through the medium; ii) \emph{sonic logs}, which give a measure of the formation’s capacity to transmit seismic waves, i.e., the formation’s interval transit time; iii) \emph{gamma ray}, which allow to derive the correlation between the radioactive isotope content and mineralogy; and iv) \emph{neutron porosity}, which tracks the average hydrogen density of the volume under investigation. Each record has a corresponding depth, and records where any of these sensor readings are missing were not considered. Consequently, every well in the data-set has at least one complete record with data from each sensor.
Besides the sensor data at each depth step, each well is identified by a string, together with its latitude and longitude. All the variables were normalised to the range $[0,1]$, with exception of the neutron porosity in which units are percentages, and are originally provided inside the specified range. Table \ref{table:sample} provides a few samples that illustrate this description.

\begin{table}[ht]
\caption{
\label{table:sample}
Data-set sample with five records. Each record includes the well Name, Latitude, and Longitude, as well as the Depth of the readings, the bulk density (RHOB), the sonic log (DT), the gamma-ray (GR), and the neutron porosity (NPHI).}
\centering
\begin{tabular}{llrrrrrrr}
\toprule
 &      Well &     Depth &      RHOB &        DT &        GR &      NPHI &       Latitude &      Longitude \\
\midrule
599151 &  F04-02-A &  0.570698 &  0.386626 &  0.185970 &  0.086413 &  0.190975 &  0.804444 &  0.512032 \\
521162 &    F16-03 &  0.721298 &  0.180663 &  0.175042 &  0.094863 &  0.434452 &  0.800649 &  0.511146 \\
320809 &    F12-01 &  0.334878 &  0.315719 &  0.319448 &  0.115855 &  0.423686 &  0.802614 &  0.513452 \\
438543 &    F02-05 &  0.203800 &  0.283569 &  0.292921 &  0.047382 &  0.254820 &  0.805221 &  0.512654 \\
103902 &    F09-03 &  0.629160 &  0.440846 &  0.193788 &  0.066738 &  0.239994 &  0.803657 &  0.513170 \\
\bottomrule
\end{tabular}
\end{table}

\begin{figure}
  \centering
  \includegraphics[scale=.5,clip=true]{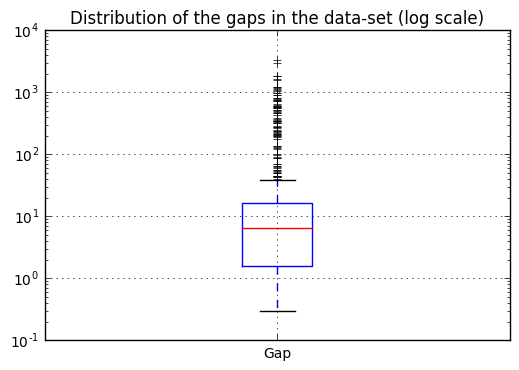}
  \qquad
  \begin{tabular}[b]{lr}
  \toprule
  {} &          Gap \\
  \midrule
  count &   650.0 \\
  mean  &    79.5 \\
  std   &   279.5 \\
  min   &     0.3 \\
  25\%   &     1.6 \\
  50\%   &     6.6 \\
  75\%   &    16.8 \\
  max   &  3361.6 \\
  \bottomrule
  \end{tabular}

  \captionlistentry[table]{Descriptive statistics for the log gaps in the data-set. The values (except the count) represent depth in meters.}
  \captionsetup{}
  \caption{
	\label{table:gapdistribution}
    Descriptive statistics for the log gaps in the data-set. The values (except the count) represent depth in meters.}
\end{figure}

\begin{figure}
\centering
\includegraphics[width=.45\textwidth]{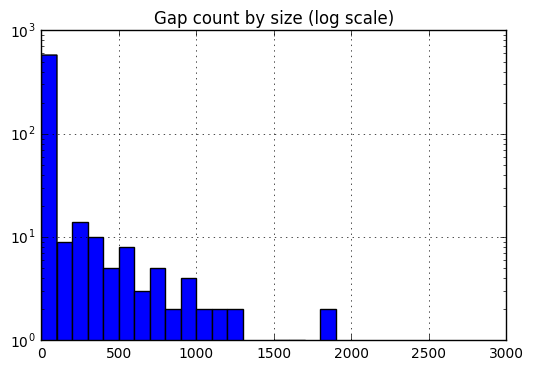}
\includegraphics[width=.45\textwidth]{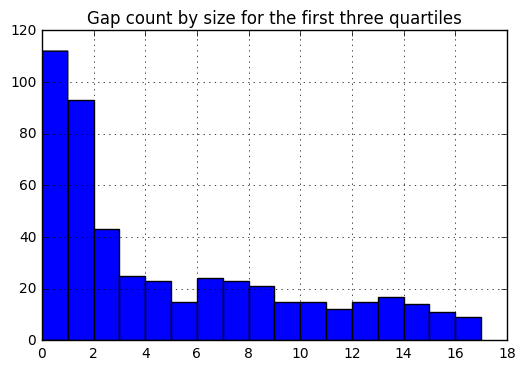}
\caption{\label{fig:gap-hist}
Left: Gap count by size for all the gaps, using the logarithmic scale; Right: Zoom over the first three quartiles, using the original scale.}
\end{figure}

In order to gain a better understanding of the problem of filling in log gaps we performed an exploratory analysis of the gaps contained in the previously described data-set. For that purpose, using the original (not normalised) depth, we considered to exist a gap in a well if the distance between two consecutive records is bigger than $0.2m$\footnote{For most of the wells the depth step is $0.1m$, but in a few cases it is $0.2m$.}. In the complete set with more than a thousand wells we identified 650 gaps from 323 different wells (the remaining wells do not exhibit gaps). Figure \ref{table:gapdistribution} shows the descriptive statistics for the gaps in the data-set, using the logarithmic scale . The values (except the count) represent depth in meters. For most of the wells each meter corresponds to ten data-points, since the depth step is $0.1$.
As one can see, most of the gaps are rather small, with $75\%$ of the cases bellow $16.8m$ ($168$ points), and the median just at $6.6m$ ($66$ points). Figure \ref{fig:gap-hist} shows the gap count by size for all the gaps (left) using the logarithmic scale, as well as a zoom over the first three quartiles (right) in the original scale.

\section{Experiments and Results}\label{sec:methodology}

Different techniques have been used to build models that predict and estimate various geophysical properties, by means of either regression or classification tasks including but not limited to: deterministic petrophysical modeling, using shale, matrix, and fluid properties; stochastic modeling, where an approximate curve is used as input, and the reconstructed curve is the output; and different soft-computing algorithms \cite{aminzadeh_neural_2006,holdaway_harness_2014}.
These approaches focus on a single well or on a few wells from the same block \cite{holmes_generating_2003, yu_method_2011,ayoub_estimating_2014,ahmadi_connectionist_2014}. 

In this work we compare the results of using Generalised Linear Models (OLS), Bayesian Regression (BRR), RANdom SAmple Consensus (RANSAC), Random Forests (RF), and Artificial Neural Networks (ANN), on the prediction of missing gaps in a single well. For that purpose  we chose a single well with complete logs (without gaps), in particular the well identified by "F02-02". From these logs we generated 30 random gaps for each gap size of the first three quartiles (respectively 16, 66, and 168 points for the first, second, and third quartiles). These will be used to average the models' performance, since it depends not only on the size of the gap but also on the values of the gap itself.
The results described in the following paragraphs were obtained with vanilla implementations of the afore mentioned algorithms, provided with the Scikit-Learn \cite{pedregosa_scikit-learn:_2011} Python machine learning library.

\begin{figure}
\centering
\includegraphics[scale=.5]{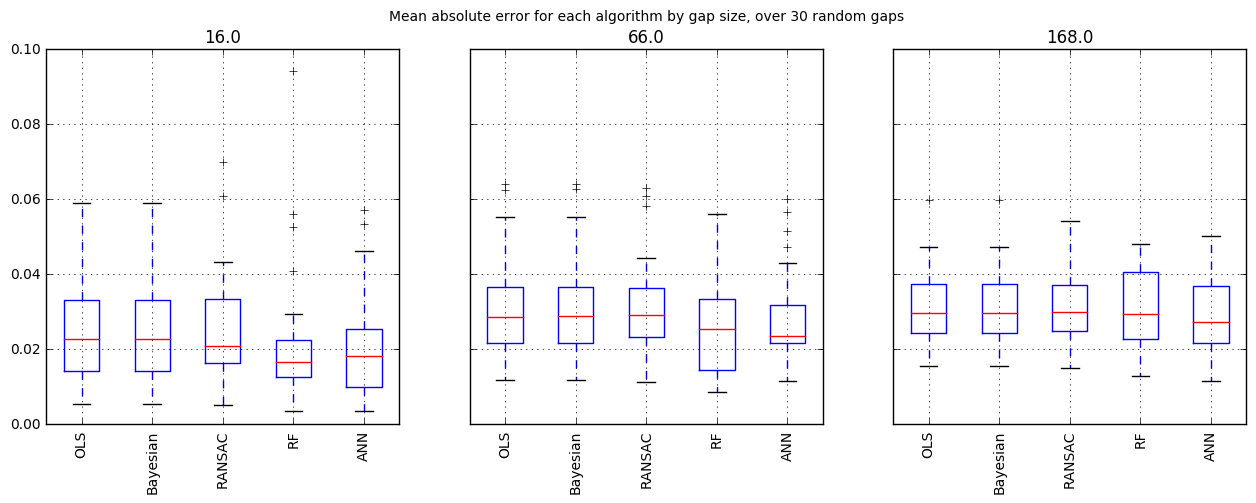}
\caption{\label{fig:r2boxplot}
Mean absolute error for each algorithm by gap size, over thirty random gaps. The values indicate percentages in the interval $\left[0, 1\right]$.}
\end{figure}

\begin{figure}
\centering
\includegraphics[scale =.55]{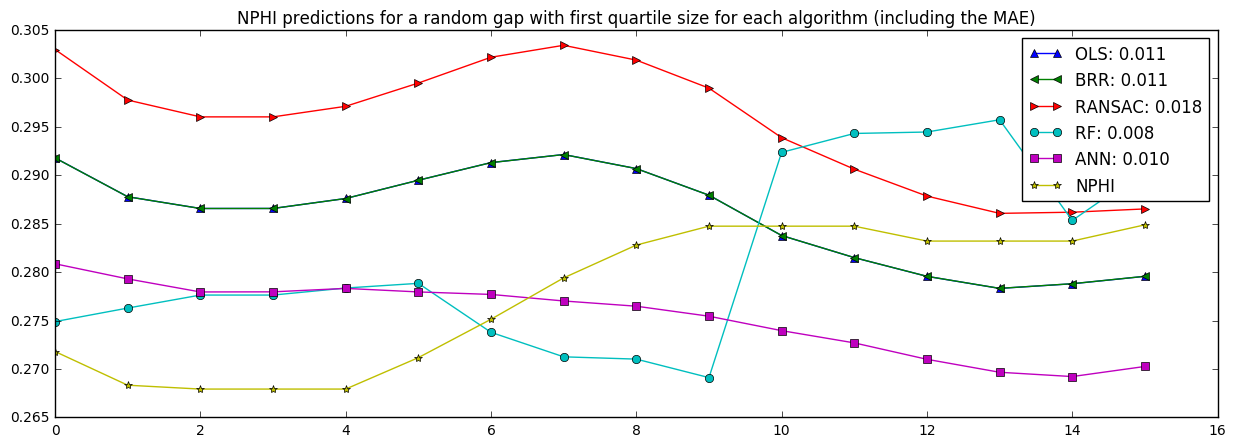}
\includegraphics[scale =.55]{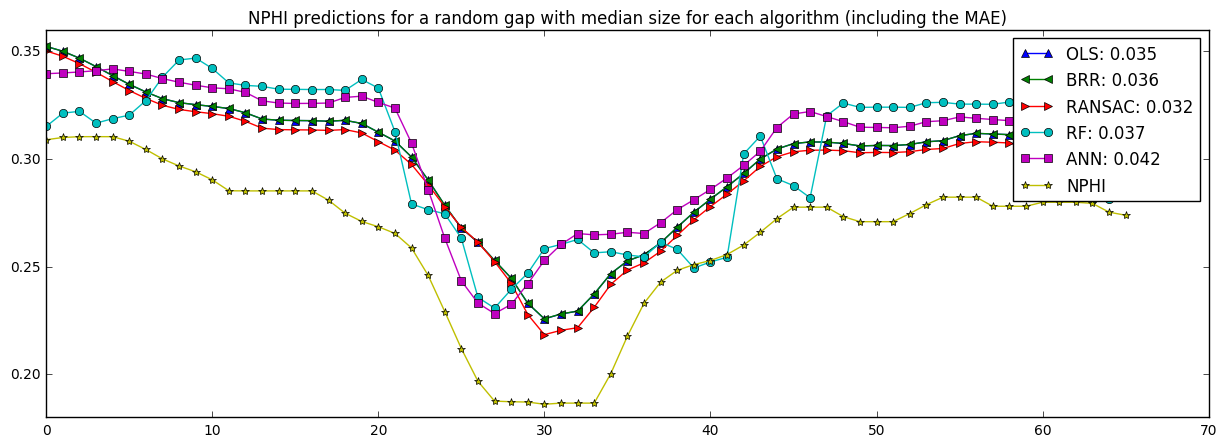}
\includegraphics[scale =.55]{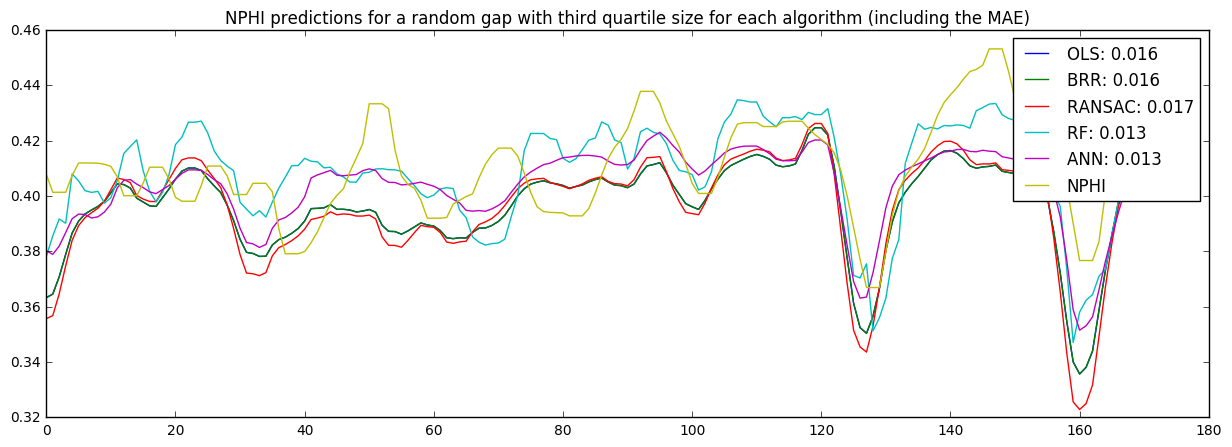}
\caption{\label{fig:mediangaps}Examples of random gaps for each gap size quartile with the corresponding predictions for each algorithm, and the mean absolute error in the legend.}
\end{figure}

The distribution of the models' errors (mean absolute error) over the thirty random gaps for each gap size can be observed in Figure \ref{fig:r2boxplot}. It shows clearly that the performance of the algorithms depends on the gap size and on the gap itself, given the variance displayed. In general the ANNs seems to perform better more often as the gap size increases. For the smaller gaps the remaining methods yield better predictions. However, the statistical difference is not significant, and both RFs and ANNs are computationally more expensive than the remaining regressions. Plotted in Figure \ref{fig:mediangaps} one can see an example gap for each of the quartiles with the corresponding predictions for each algorithm, and the mean absolute error in the legend (eg., OLS: $0.01$).

\section{Future Work}\label{sec:future}

The goal of this study was first to gain a better understanding of the gaps across a big set of wells, and second to build a data-driven model for filling gaps in sensor data from well logs, using the remaining logs as features, at any depth of a particular well.

The analysis of the gap sizes has shown that in most cases the gaps are rather small, especially when compared to the complete depth of the wells. Nevertheless, predicting values for a gap in a sensor log from the remaining sensor logs is not an easy task as the results demonstrated. The smaller the gap, the more obvious is the difference between the observations and the models, even though the mean absolute error is slightly lower. The experiments have shown as well that the algorithms perform differently depending on the gap values, not only on the gap size.

Future work can delve into: i) collecting more logs, and re-analyse the gaps on a log-by-log basis; ii) fine-tuning each algorithm; iii) cluster samples by log similarity and develop different models for the same well; iv) increase the number of random gaps used to evaluate the algorithms; v) add more algorithms to the analysis. These are only a few of the immediate possibilities to foster a deeper understanding of the well log gaps and how to fill them in. Other approaches such as feature engineering and extraction are also a mandatory avenue of research in order to improve the prediction performance.

\vfill
\bibliographystyle{alpha}

\begin{thebibliography}{MEHA15}

\bibitem[AAHE14]{ahmadi_connectionist_2014}
Mohammad-Ali Ahmadi, Mohammad~Reza Ahmadi, Seyed~Moein Hosseini, and Mohammad
  Ebadi.
\newblock Connectionist model predicts the porosity and permeability of
  petroleum reservoirs by means of petro-physical logs: {Application} of
  artificial intelligence.
\newblock {\em Journal of Petroleum Science and Engineering}, 123:183--200,
  2014.

\bibitem[ABB14]{aifa_neuro-fuzzy_2014}
Tahar Aifa, Rafik Baouche, and Kamel Baddari.
\newblock Neuro-fuzzy system to predict permeability and porosity from well log
  data: {A} case study of {Hassi} {R}׳ {Mel} gas field, {Algeria}.
\newblock {\em Journal of Petroleum Science and Engineering}, 123:217--229,
  2014.

\bibitem[ADG06]{aminzadeh_neural_2006}
F.~Aminzadeh and P.~De~Groot.
\newblock {\em Neural networks and other soft computing techniques with
  applications in the oil industry}.
\newblock EAGE Publications, 2006.

\bibitem[AM14]{ayoub_estimating_2014}
Mohammed~Abdallah Ayoub and Ahmed~Abdelhafeez Mohamed.
\newblock Estimating the {Lengthy} {Missing} {Log} {Interval} {Using} {Group}
  {Method} of {Data} {Handling} ({GMDH}) {Technique}.
\newblock {\em Applied Mechanics and Materials}, 695:850, 2014.

\bibitem[HHH03]{holmes_generating_2003}
Michael Holmes, Dominic Holmes, and Antony Holmes.
\newblock Generating {Missing} {Logs} {B} {Techniques} and {Pitfalls}.
\newblock In {\em {AAPG} {Annual} {Meeting}, {Salt} {Lake} {City}, {Utah}},
  2003.

\bibitem[Hol14]{holdaway_harness_2014}
Keith Holdaway.
\newblock {\em Harness {Oil} and {Gas} {Big} {Data} with {Analytics}:
  {Optimize} {Exploration} and {Production} with {Data} {Driven} {Models}}.
\newblock Wiley Publishing, 1st edition, 2014.
\newblock bibtex: Holdaway:2014.

\bibitem[MEHA15]{mohseni_application_2015}
Hassan Mohseni, Moosa Esfandyari, and Elham Habibi~Asl.
\newblock Application of artificial neural networks for the prediction of
  carbonate lithofacies, based on well log data, {Sarvak} {Formation}, {Marun}
  oil field, {SW} {Iran}.
\newblock {\em Geopersia}, 5(2):111--123, 2015.

\bibitem[PVG{\etalchar{+}}11]{pedregosa_scikit-learn:_2011}
F.~Pedregosa, G.~Varoquaux, A.~Gramfort, V.~Michel, B.~Thirion, O.~Grisel,
  M.~Blondel, P.~Prettenhofer, R.~Weiss, V.~Dubourg, J.~Vanderplas, A.~Passos,
  D.~Cournapeau, M.~Brucher, M.~Perrot, and E.~Duchesnay.
\newblock Scikit-learn: {Machine} {Learning} in {Python}.
\newblock {\em Journal of Machine Learning Research}, 12:2825--2830, 2011.
\newblock bibtex: scikit-learn.

\bibitem[YSM11]{yu_method_2011}
Y.~Yu, D.J. Seyler, and M.D. McCormack.
\newblock {\em Method for estimating missing well log data}.
\newblock Google Patents, November 2011.
\newblock US Patent 8,065,086.

\bibitem[ZAB14]{zerrouki_prediction_2014}
Ahmed~Ali Zerrouki, Tahar A{\textbackslash}"{\textbackslash}ıfa, and Kamel
  Baddari.
\newblock Prediction of natural fracture porosity from well log data by means
  of fuzzy ranking and an artificial neural network in {Hassi} {Messaoud} oil
  field, {Algeria}.
\newblock {\em Journal of Petroleum Science and Engineering}, 115:78--89, 2014.

\end{thebibliography}

\newcommand{\etalchar}[1]{$^{#1}$}

\end{document}